\def\year{2019}\relax
\documentclass[letterpaper]{article} %DO NOT CHANGE THIS
\usepackage{ijcai18}  %Required

\usepackage[utf8]{inputenc} % allow utf-8 input
\usepackage[T1]{fontenc}    % use 8-bit T1 fonts
\usepackage{hyperref}       % hyperlinks
\usepackage{url}            % simple URL typesetting
\usepackage{booktabs}       % professional-quality tables
\usepackage{amsfonts}       % blackboard math symbols
\usepackage{nicefrac}       % compact symbols for 1/2, etc.
\usepackage{microtype}      % microtypography

\usepackage[dvipdfmx]{graphicx}
\usepackage{mediabb}

\usepackage{here}

\usepackage{times}
\usepackage{soul}
\usepackage[small]{caption}

\usepackage{color}
\usepackage{xcolor}
\usepackage{amsmath}
\usepackage{amsfonts}
\usepackage[subrefformat=parens]{subcaption}

\title{Dual Convolutional Neural Network for Graph of Graphs Link Prediction\\
\vspace{4mm}
\Large{
Shonosuke Harada$^1$,
Hirotaka Akita$^1$,
Masashi Tsubaki$^2$,
Yukino Baba$^{3,6}$, \\
Ichigaku Takigawa$^4$,
Yoshihiro Yamanishi$^5$,
Hisashi Kashima$^{1,6}$}\\
\vspace{3mm}
\small{
$^1$ Kyoto University \\
$^2$ National Institute of Advanced Industrial Science and Technology\\
$^3$ University of Tsukuba \\
$^4$ Hokkaido University \\
$^5$ Kyushu Instutute of Technology  \\
$^6$ RIKEN Center for Advanced Intelligence Project  
}
}

\newcommand{\Tsubaki}[1]{\textcolor{blue}{[Tsubaki] #1}}

\begin{document}
\maketitle
\begin{abstract}
Graphs are general and powerful data representations which can model complex real-world phenomena, ranging from chemical compounds to social networks; however, effective feature extraction from graphs is not a trivial task, and much work has been done in the field of machine learning and data mining. The recent advances in graph neural networks have made automatic and flexible feature extraction from graphs possible and have improved the predictive performance significantly. In this paper, we go further with this line of research and address a more general problem of learning with a {\it graph of graphs (GoG)} consisting of an external graph and internal graphs, where each node in the external graph has an internal graph structure. We propose a dual convolutional neural network that extracts node representations by combining the external and internal graph structures in an end-to-end manner. Experiments on link prediction tasks using several chemical network datasets demonstrate the effectiveness of the proposed method.
\end{abstract}

\section{Introduction}
Graphs are general and powerful data representations which can model complex real-world phenomena,
ranging from chemical compounds to social networks.
However, most of the existing data analysis techniques assume that
each data instance is readily represented as a fixed-dimensional feature vector;
therefore, graph-structured data analytics has been one of the topics
fascinating researchers in the field of machine learning and data mining.

In chemoinformatics, chemical compounds are often represented as molecular graphs
whose nodes correspond to their atoms and edges correspond to the chemical bonds among them.
Molecular fingerprinting~\cite{morgan1965generation} is a widely used way for molecular graph representation
that uses a set of subgraphs responsible for important chemical properties.
A molecular fingerprint is a fixed-dimensional binary vector,
each of whose elements corresponds to a subgraph (e.g., benzene ring)
related to some chemical property (e.g., aromatic).
Machine learning methods have been successfully applied to prediction of
various molecular properties such as drug efficacy~\cite{gamo2010thousands}.

\begin{figure*}[tb]
 \begin{minipage}[b]{0.4\hsize}
  \begin{center}
   \includegraphics[width=\linewidth]{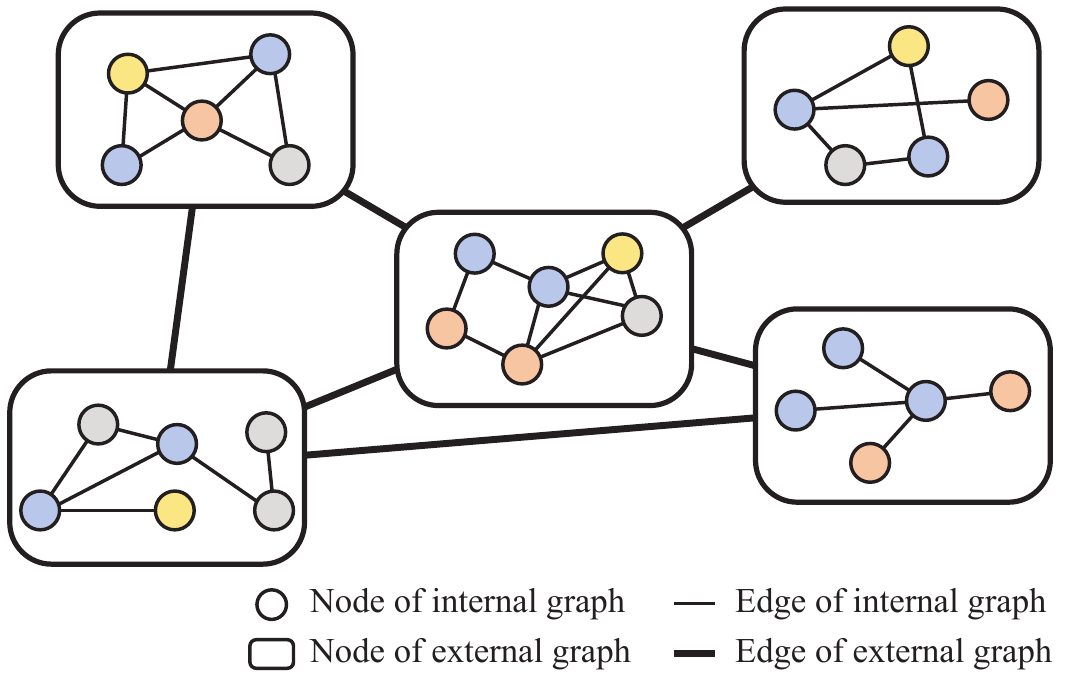}
  \end{center}
  \vspace{2cm}
  \caption{A (two-level) GoG consists of an external graph and internal graphs.
  In a drug interaction network, for example, each node of the external graph corresponds to a drug molecule,
  and each molecule has its own internal graph structure representing chemical bonds among its atoms.}
  \label{fig:graphofgraph}
 \end{minipage}
 \hspace{2mm}
 \begin{minipage}[b]{0.6\hsize}
 \begin{center}
  \includegraphics[width=\linewidth]{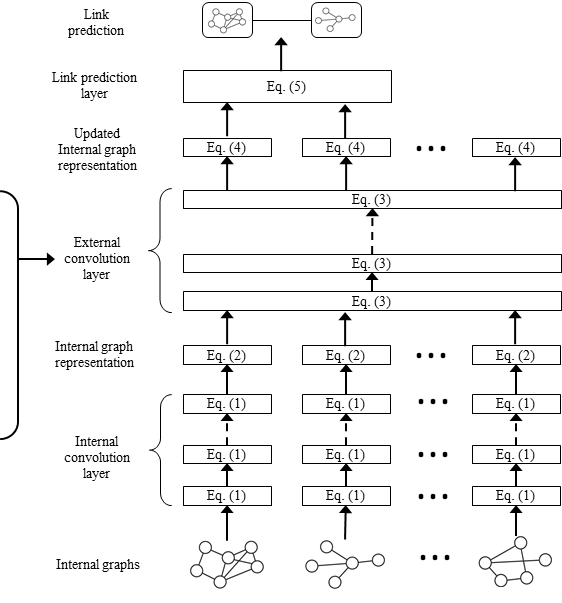}
 \end{center}
  \caption{The dual convolution architecture for a GoG. The internal convolution layer extracts features from the internal graphs, which are followed by the external convolutions layer to incorporate structural information of the external graph.}
  \label{fig:architecture}
 \end{minipage}
\end{figure*}

\iffalse
\begin{figure}[tb]
  \begin{center}
  \includegraphics[width=100mm]{images/gog_colored.pdf}
  \end{center}
  \caption{
  A two-layer GoG consists of an external graph and internal graphs.
  In a drug interaction network, for example, each node of the external graph corresponds to a drug molecule,
  and each molecule has its own internal graph structure representing chemical bonds among its atoms.
  Generally, a GoG can have more than two layers.
  In this paper, we only consider a two-layer GoGs for simplicity,
  our idea is easily generalized to a GoGs with more layers.
  }
  \label{fig:graphofgraph}
\end{figure}
\fi

Aside from the graph structures {\it inside} molecules,
other types of graphs such as various interaction networks {\it among} molecules are sometimes available,
where the nodes correspond to the molecules and the links corresponds to the chemical interactions between them.
This hierarchically structured graph has two types of graph structures:
the internal graph structure inside a single molecule and the external graph structure among a set of molecules,
which is known as a \emph{graph of graphs} (GoG) (Fig.~\ref{fig:graphofgraph}).

Our focus in this paper is to develop an effective modeling method for the GoG
which has a more general and complex graph structure than a single graph,
and to consider the link prediction task on a GoG.
For example, predicting links in a chemical interaction network provides useful information
for drug repositioning (to find new applications of existing drugs)
and finding their potential side effects~\cite{lounkine2012large,medina2013shifting}.

Recently, deep neural networks (DNNs) have been successfully applied to graph-structured data.
Many of the existing approaches employ recursive constructions of graph representations
called {\it graph convolution}~\cite{NIPS2015_5954} and
end-to-end learning of the whole networks using backpropagation~\cite{sutskever2014sequence}.
Such representation learning has an advantage over existing approaches
based on off-the-shelf features such as the molecular fingerprints,
since it enables automatic and flexible feature extraction from graph
and improves the predictive performance.

% In this paper, we extend the graph convolution approach to GoG
% by introducing a new neural network architecture called \emph{dual convolution}
% that seamlessly handles both internal and external graph structures in an end-to-end manner
% to learn low-dimensional representations of the GoG nodes.
% Our experiments of the link prediction task using three real GoG datasets demonstrate
% that the dual convolutional neural network achieves a superior performance to baseline methods
% depending on the network density, degree distribution, and dataset size.

% イントロ最終段落は、こんな感じにしてみました。
In this paper, we extend the graph convolutional neural network to GoG
by introducing a new architecture called \emph{dual convolution}.
The dual convolution allows us to
(i)~seamlessly handle both internal and external graph structures in an end-to-end manner
and (ii)~efficiently learn low-dimensional representations of the GoG nodes.
We conduct experiments of the link prediction task using real GoG datasets such as a drug-drug interaction network, and show that the dual convolution achieves a superior performance to baselines in some datasets.
% \kashima{この辺からちょっとよくわからない}
In addition, we analyze the effects of network density and degree distribution for performance;
indeed, real-world external graphs (e.g., sparse and heavy-tailed networks)
have a variety of structures compared to internal graphs (e.g., small organic molecules).
We believe that our performance analysis of dual convolution
provides insights into developing practical applications for real-world datasets.

\section{Link Prediction Problem in a Graph of Graphs (GoG)}

Throughout the paper, we denote vectors by bold lowercase letters
(e.g., $\mathbf v \in \mathbb R^d$), matrices by bold uppercase letters
(e.g., $\mathbf M \in \mathbb R^{m \times n}$), and
scalars and discrete symbols (such as graphs and nodes) by non-bold letters
(e.g., $\mathcal G$ and $n$).

A GoG is a hierarchically structured graph $\mathcal G = (\mathcal V, \mathcal A)$,
where $\mathcal V$ is the set of nodes, $\mathcal A$ is the adjacency list.
Each node in the GoG is also a graph, which 
%Then we consider $G_i \in \mathcal V$, which is the $i$-th node in $\mathcal G$.
%For simplicity, we denote $G_i = G$ in the following description.
%A node of a GoG, $G$, is also a graph, i.e., 
we denote by $G = (V, A) \in \mathcal V$,
where $V$ is the set of nodes, and $A$ is the adjacency list.
We refer to $\mathcal G$ as an {\it external graph} and $G$ as an {\it internal graph}\footnote{Generally, a GoG can have more than two levels. In this paper, we only consider two levels for simplicity, and refer to them by internal graph and external graph; however, our fundamental idea itself is easily generalized to GoGs with more levels.}.

For example, an interaction network between chemical compounds is represented as a GoG $\mathcal G$, whose nodes $\mathcal V$ are the set of compounds, and whose edges referred to by its adjacency list $\mathcal A$ are the set of binary relations (e.g., interact or not) among the compounds.
For each compound $G = (V, A) \in \mathcal V$, $V$ is the set of the atoms included in the compound, and $A$ indicates the set of chemical bonds among the atoms.
%For simplicity, we do not consider the chemical bonds $E$ in molecular graphs
%and use only atoms and their adjacency, i.e., $G = (V, A)$ instead of $G = (V, A, E)$,
%which is similar to the work of \cite{NIPS2015_5954}.
%We refer to $\mathcal G$ as an ``external graph'' and $G$ as an ``internal graph''.
%In Section 3, we propose the use of graph convolution \citep{NIPS2015_5954} for both the external and the internal graphs defined above.

Given a GoG, our goal is to obtain a feature representation of each internal graph $G \in \mathcal V$ and to predict the probability of the existence of a (hidden) link between arbitrary two internal graphs $G_i, G_j \in \mathcal V$.
\section{Proposed Method: Dual Convolution}

We propose the {\it dual convolutional neural network} for a GoG that consists of three components (Fig.~\ref{fig:architecture}): the internal graph convolution layer (Section~3.1), the external graph convolution layer (Section~3.2), and the link prediction layer (Section~3.3). 

\iffalse
\begin{figure}[tb]
  \begin{center}
   \includegraphics[width=100mm]{images/architecture.pdf}
  \end{center}
  \caption{
  Dual convolution architecture for a GoG.
  }
  \label{fig:architecture}
\end{figure}
\fi

\subsection{Internal graph convolution}

The internal convolution layer takes an internal graph $G = (V, A)$ (e.g., a chemical compound) as its input and gives a fixed-dimensional vector representation for the graph.
At the bottom of the internal convolution layer,
%As with the graph convolution of \cite{NIPS2015_5954},
the low-dimensional real-valued vector representation $\mathbf v_k \in \mathbb R^d$ for the $k$-th node $v_k \in V$ is randomly initialized, where $d$ is the dimension of the vector.
%Note that the dimensionality is one of the model hyperparameters,
Each $\mathbf v_k$ is initialized differently depending on the types of nodes (e.g., hydrogen or oxygen), and trained using backpropagation as well as the subsequent external convolution and link prediction layers in an end-to-end manner  (Section~3.3).

Given the initialized node feature $\mathbf v_k$ for each node $v_k$,
starting from $\mathbf v_k^{(0)} = \mathbf v_k$,
%we denote $\mathbf v_k$ at time step $t$ as $\mathbf v_k^{(t)}$
we update ${\mathbf v}_k^{(t)}$ to ${\mathbf v}_k^{(t+1)}$ by the {\it internal convolution} operation:
\begin{eqnarray}
\label{internal_convolution}
\mathbf v_k^{(t+1)} = f_G \left( \mathbf W \mathbf v_k^{(t)}
+ \sum_{v_m \in A_k} \mathbf M \mathbf v_m^{(t)} \right),
\end{eqnarray}
where $f_G$ is the non-linear activation function such as $\mbox{ReLU}$\footnote{We have freedom of choice in our model such as the non-linear activation function $f_G$; we give their specifications in Section 4.2.},
$A_k$ is the adjacency list of $v_k$, and $\mathbf W \in \mathbb R^{d \times d}$ and $\mathbf M \in \mathbb R^{d \times d}$ are the weight matrices to be learned.
As with the graph convolution of Duvenaud {\it et al.}~\cite{NIPS2015_5954}, 
each node gradually incorporate global information on the graph into its representation by iterating the internal convolution step using the representations of its adjacent nodes.
We make $T$ iterations to obtain 
%${\mathbf v}_k^{(T)}$.
${\mathbf v}_k^{(1)}, {\mathbf v}_k^{(2)}, \dots , {\mathbf v}_k^{(T)}$.
%Section~4.2 also describes the implementation details of other equations in this section.

Finally, summing all of the node features over all of the internal convolution steps to obtain the internal graph representation as 
\begin{eqnarray}
\label{internal_output}
\mathbf g^{(T)} = \sum_{v_k \in V} \sigma_G \left( \sum_{t=0}^{T} \mathbf v_k^{(t)}\right),
\end{eqnarray}
where $\sigma_G$ is a non-linear function such as the softmax function.
%\update{The summation over all time stamps is to incorporate the information of various ranges.}

In the following, we denote by $\mathbf g_i^{(T)}$ the representation  of internal graph $G_i \in \mathcal V$, which will be the initial feature vector in the external graph convolution introduced in the next section.

\subsection{External graph convolution}
The set of representations for all the internal graphs $\{ \mathbf g_i^{(T)} \}_{G_i \in {\mathcal V}}$ are further updated with the {\it external convolution} to incorporate structural information of the external graph.
Starting from $\ell=0$, we make $L$ updates using the external convolution operation given as 
%Given the above internal graph representation $\mathbf g_i^{(0)}$ and external graph $\mathcal G = (\mathcal V, \mathcal A, \mathcal E)$,
%we update the feature vector using a similar convolution function as Eq.~(\ref{internal_convolution}) as follows:
\begin{eqnarray}
\label{external_convolution}
\mathbf g_i^{(T+\ell+1)} = f_{\mathcal G} \left(
\mathbf U \mathbf g_i^{(T+\ell)} + \sum_{G_m \in \mathcal A_i}
\mathbf V \mathbf g_{m}^{(T+\ell)} \right),
\end{eqnarray}
where $f_{\mathcal G} $ is a non-linear function,
$\mathcal A_i$ is the adjacency list of internal graph $G_i$ in the external graph,
and $\mathbf U \in \mathbb R^{d \times d}$ and $\mathbf V \in \mathbb R^{d \times d}$
are the weight matrices to be learned.

Finally, we obtain the final internal graph representation ${\mathbf h}_i^{(T+L)}$ considering all of the $L$ external convolution steps as
\begin{eqnarray}
\label{external_output}
\mathbf h_i^{(T+L)} = \sigma_{\mathcal G} \left(\sum_{\ell=0}^L \mathbf g_i^{(T+\ell)}\right),
\end{eqnarray}
where $\sigma_{\mathcal G}$ is a  non-linear function such as the softmax function.
Note that our dual convolution does not aim to obtain a single representation of the external graph, but to obtain the representation of each internal graph considering both the internal and external graphs, which will be used in the following link prediction layer.
%$\mathbf g_i^{(L)}$ is the representation of internal graph $G_i$ at time step $L$,
%i.e., our dual convolution does not aim to obtain the representation of external graph $\mathcal G$.
%In this paper, we call $\mathbf g_i^{(L)}$ the dual convolutional representation of an internal graph,
%which is the input feature vector for link prediction on an external graph, described in the next section.

\subsection{End-to-end learning of the link prediction function}

The link between two internal graphs $G_i$ and $G_j$ is predicted
using their final representations $\mathbf h_i^{(T+L)}$ and $\mathbf h_j^{(T+L)}$.
A multi-layer neural network $p$ outputs a two-dimensional vector $\mathbf y \in \mathbb R^2$:
\begin{eqnarray}
\label{neural_network}
\mathbf y = p \left( \mathbf h_i^{(T+L)}, \mathbf h_j^{(T+L)} \right),
\end{eqnarray}
and the softmax function gives the final link probability:
%\begin{equation}
$p_{t} =  {\exp(y_t)} / {\sum_k \exp(y_k)}$,
%\nonumber
%\end{equation}
where $t \in \{ 0, 1 \}$ is the binary label (i.e., link or no-link). % and $p_t$ is the probability of $t$.
Note that the symmetry of $p$ with respect to its two inputs is ensured by its specific implementation described in Section 4.2.

%\subsection{Optimization}

Given a set of all internal graphs and some links among them as the training dataset,
we minimize the cross-entropy loss function:
$\mathcal L(\Theta) = -\sum_{i=1}^{N} \log p_{t_i}$
with respect to the model parameters $\Theta$ including the set of all weight matrices in the dual convolution network
and the atom features (that are initialized randomly).
$N$ is the total number of internal graph pairs in the training dataset,
$t_i$ is the $i$-th label (link or no-link).

\section{Experiments}

We evaluate the idea of the dual convolution that combines the structural information of both internal and external graphs in a GoG.
We compare the link prediction accuracy of the proposed method and several baselines using three real chemical networks.
Overall, the proposed method works well for moderately dense external networks with heavy-tailed degree distributions. 
In an extremely sparse and light-tailed external network, external links are almost useless, and the domain specific features (Morgan indices) perform the best. 
The internal convolution also suffers from the lack of external links as the training data. 

%In networks with heavy-tailed degree distributions, external links are most informative; at one extreme, only external links are enough for making good predictions (Figure 3).

% \begin{figure*}[tbh]
%  \begin{minipage}{0.5\hsize}
%   \begin{center}
%    \includegraphics[bb= 0 0 480 540, width=90mm]{images/minidpdegree.pdf}
%   \end{center}
%   \subcaption{Drug-Protein}
%   \label{fig:one}
%  \end{minipage}
%  \begin{minipage}{0.5\hsize}
%  \begin{center}
%   \includegraphics[bb= 0 0 480 540, width=90mm]{images/minicidegree.pdf}
%  \end{center}
%   \subcaption{Chemical-Interaction}
%   \label{fig:three}
%  \end{minipage}
%  \caption{XXX}
% \end{figure*}

\begin{figure*}[tb]
\captionsetup[subfigure]{justification=centering}
 \begin{minipage}{0.33\hsize}
  \centering
   \includegraphics[width=55mm]{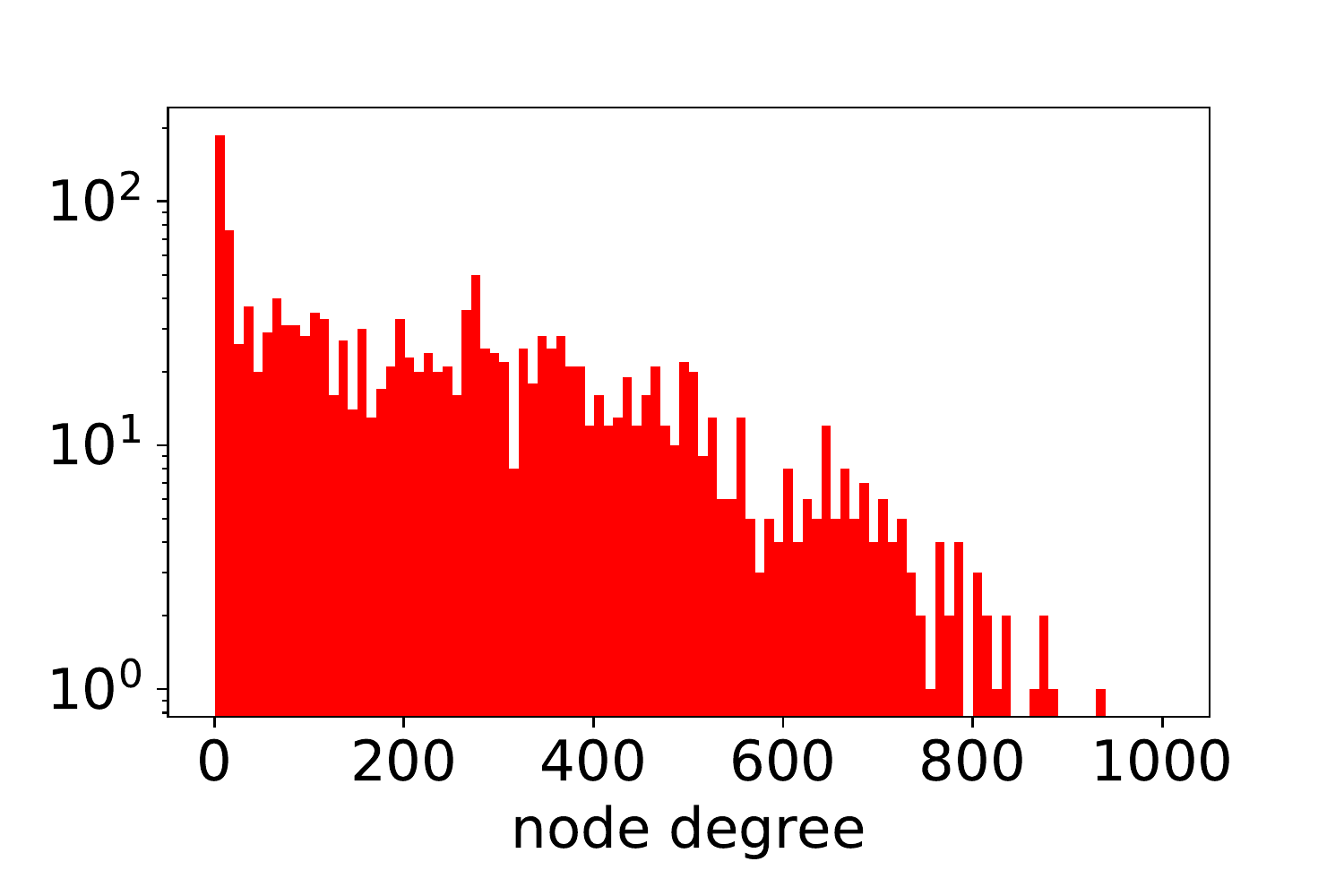}
  \subcaption{Drug-drug interaction \\ (dense, extremely heavy-tailed)}
  \label{fig:one}
 \end{minipage}
 \begin{minipage}{0.33\hsize}
 \begin{center}
  \includegraphics[width=55mm]{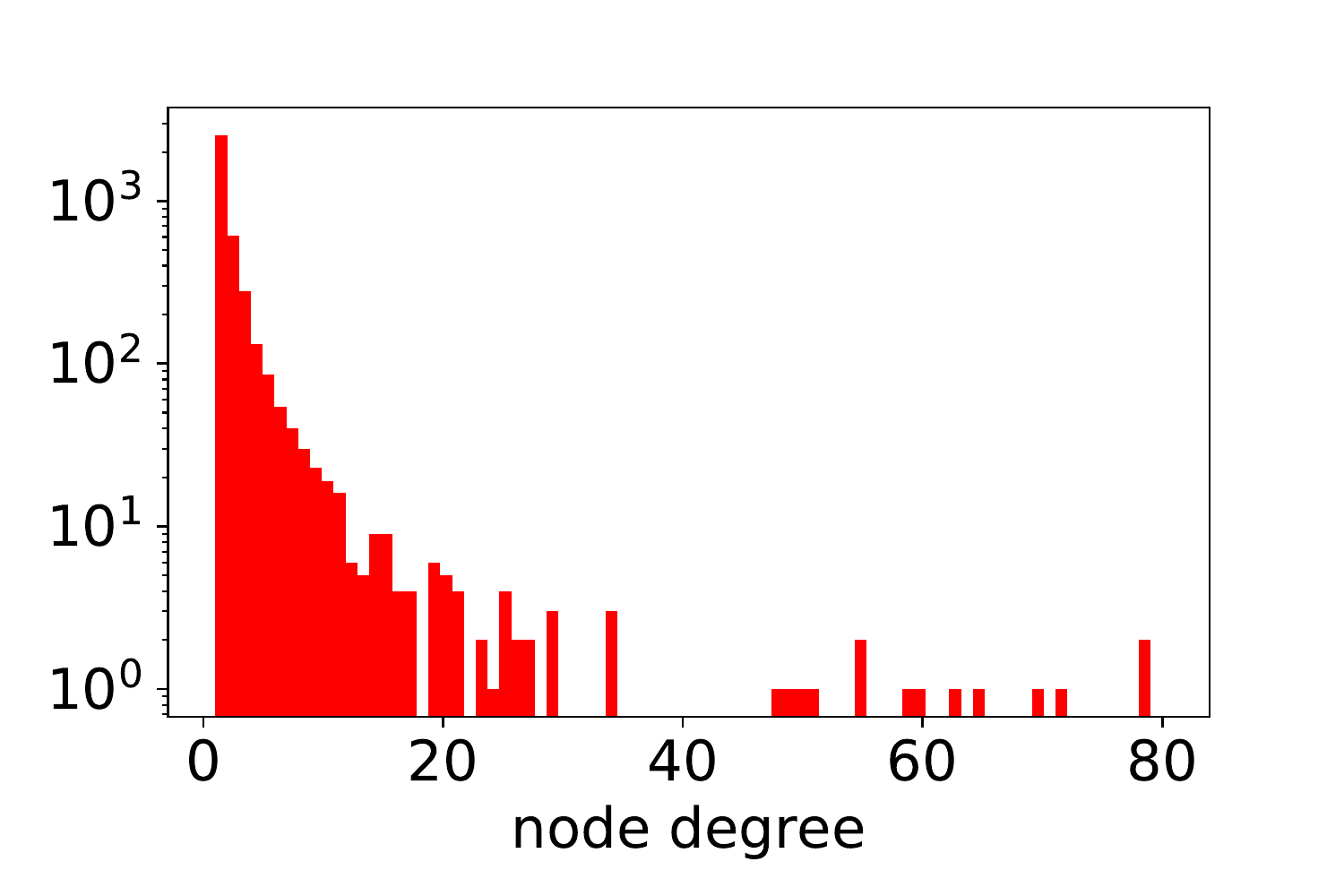}
 \end{center}
  \subcaption{Drug-function \\ (sparse, heavy-tailed)}
  \label{fig:two}
 \end{minipage}
 \begin{minipage}{0.33\hsize}
 \begin{center}
  \includegraphics[width=55mm]{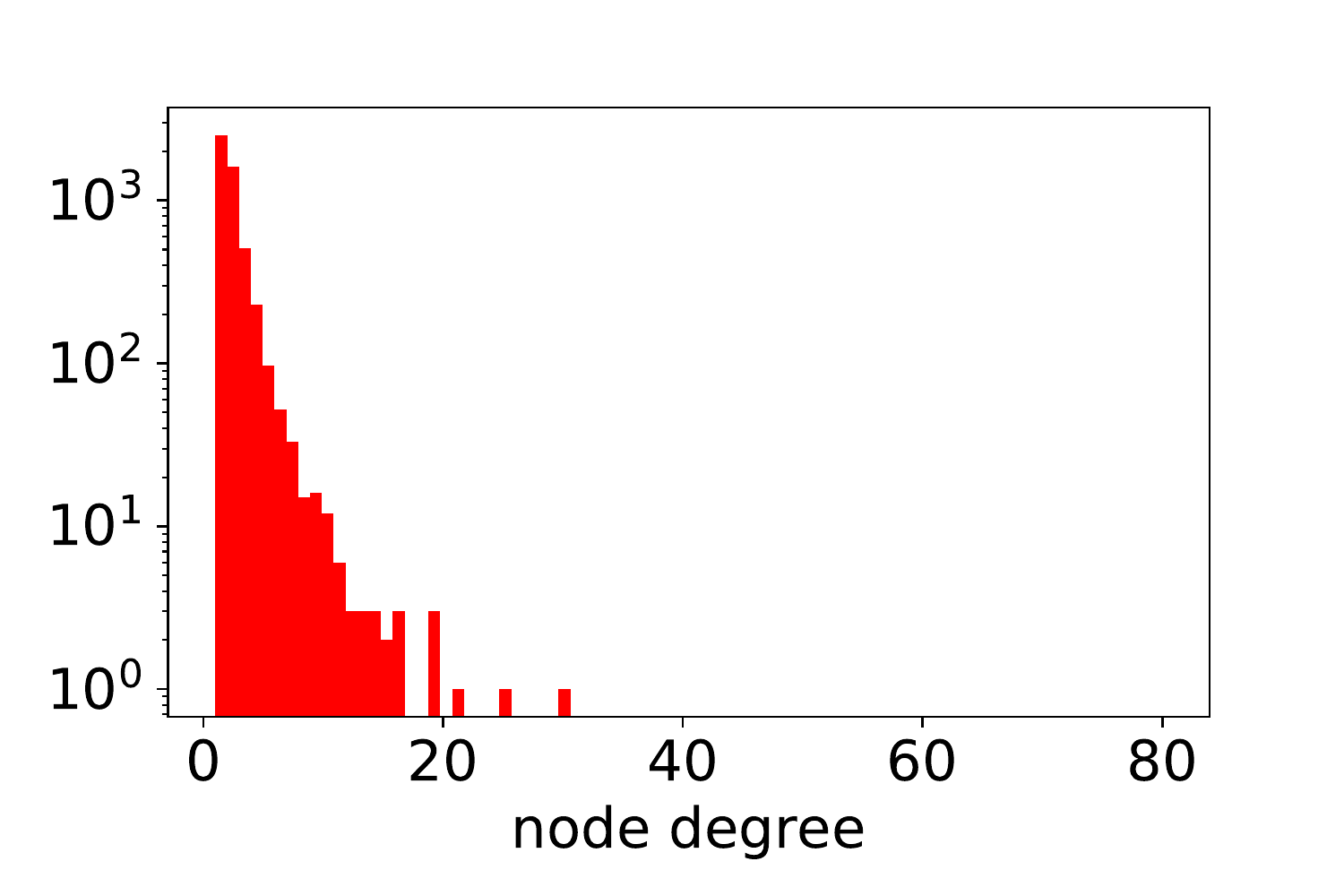}
 \end{center}
  \subcaption{Metabolite \\ (extremely sparse, light-tailed)}
  \label{fig:three}
 \end{minipage}
 \caption{\small{Node degree distributions of the external graphs of the three GoG datasets.
% The nodes with more than ten neighborhoods are removed because the degrees of only a few nodes are more than ten in the drug function dataset and in the metabolite dataset.
%各データの次数分布。Drug-InteractionとMetaboliteについては次数が大きいノードはほとんど存在していないので、次数10以下の分布のみを表示している。
\label{fig:degree}
 }}
\end{figure*}

\subsection{Specific implementation of the proposed model}
We have freedom of choice in our dual convolutional network such as the nonlinear activation functions and the depth of each layer.
In the experiments, we use the following specific choices.
In the internal graph convolution (\ref{internal_convolution}), we use $\mbox{ReLU}$ as function $f_G$, and use different $\mathbf W$ and $\mathbf M$ for different degrees ($|A_k|$ and $|A_m|$) and convolutional steps.
%Note that the consideration of degree in weight matrices
%is based on the graph convolution by Duvenaud {\it et al.} (\citeyear{NIPS2015_5954}).
%\update{We ignored the chemical bond types mainly for computational efficiency; the data size is increased by encoding the bond information as adjacency matrices. This is compensated to some extent by introducing the different parameter matrices for different node degrees by following Duvenaud {\it et al.} (\citeyear{NIPS2015_5954}).}
The node representation $\mathbf{v}_{k}$ is randomly initialized depending on their atom types, e.g., $\mathbf{v}_{\text{hydrogen}} = [0.1, 0.5]^\top$ and $\mathbf{v}_{\text{oxygen}} = [-0.4, 1.2]^\top$ (if $d=2$) which are generated from a same Gaussian distribution.
We set the number of dimension of the internal graph representations as $d=64$.
%
% \\\\
% {\bf\ \kashima{なんて言う？} (Eqs.~(\ref{internal_output}) and (\ref{external_output}))}:
% we use $\mbox{softmax}$ as function $g$.
% \\\\
%
When we obtain the internal graph representations, ${\mathbf g}^{(T)}$ in the internal convolution (\ref{internal_output}) and ${\mathbf h}^{(T+L)}$ in the external convolution (\ref{external_output}), we use the $\mbox{softmax}$ function as $\sigma_G$ and $\sigma_{\mathcal G}$.

In the external graph convolution (\ref{external_convolution}), we use the $\mbox{softmax}$ function as $f_{\mathcal G}$, and use different $\mathbf U$ and $\mathbf V$ for different convolutional steps;
we do not distinguish different degrees because the interaction networks have larger numbers of degrees than molecular graphs as shown in Fig.~\ref{fig:degree}.

We use the two-layer neural network as the link prediction network~(\ref{neural_network}) whose input is given as $\left( \mathbf h_i^{(T+L)}+ \mathbf h_j^{(T+L)} \right)\oplus\left(\mathbf h_i^{(T+L)} \odot  \mathbf h_j^{(T+L)}\right)$,
where $\oplus$ is the concatenation of two vectors and $\odot$ is the Hadamard product. Note that the input is symmetric with respect to $\mathbf h_i^{(T+L)}$ and $\mathbf h_j^{(T+L)}$.
The layer sizes are $(128, 64, 2)$, and all the non-linear activation functions are $\mbox{ReLU}$.

%
%We randomly chose $1{,}000$ positive links and $1{,}000$ negative links that were not included in the training dataset, and used them for \update{sampling approximations of the whole imbalanced set.}

%\update{Some papers (e.g. “Semi-supervised Penalized Output Kernel Regression for Link Prediction” (ICML 2011) and “Evaluating Link Prediction Methods” (KAIS 45(3)) use or sometimes recommend AUC-PR (Precision-Recall curve). So we also evaluated PR-AUC.}

%We only used molecules whose numbers of atoms are fewer than $64$ in our experiments.
%We used ROC-AUC for evaluating the prediction performance.

We implement the proposed dual convolution using Chainer~\cite{tokui2015chainer} and use ADAM~\cite{kingma2014adam} as the optimizer.
% The batch size is set to $256$ when the number of training data $2n$ is greater than $10{,}000$, and is set to $128$ when the number of training data is less than $5{,}000$ in the drug--function dataset and in the drug--drug dataset.
% It is set to $64$ in the metabolite dataset.
The batch size is set to $256$  in the drug-drug dataset and in the drug-function dataset, and set to $128$ in the metabolite dataset.
% The batch size was set to $256$ in the drug--function dataset, $128$ in the drug--drug dataset, and $64$ in the metabolite dataset.
% The numbers of convolution steps $T$ and $L$ are tuned using the grid search; the candidate values are $\{1, 3, 5\}$ for the drug--function dataset, $\{3, 5, 8\}$ for the drug--drug dataset, and $\{1, 3\}$ for the metabolite dataset.
% % \baba{grid searchしているのは$T$だけ？$L$も？}
% % \baba{over the validation set?}
% We also set the dropout rate from $\{0.0, 0.2, 0.5\}$.
% We use a validation set to tune those hyperparameters\footnote{Due to the limitation of the computational resources, we could not help limiting the hyperparameter ranges rather ``intuitively'' depending on the network sparsity and other data characteristics to perform thorough search over all possible hyperparameter combinations.}.
The numbers of convolution steps $T$ and $L$ are tuned using the grid search; the candidate values are $\{1, 3, 5\}$ for all the dataset except $T$ for the metabolite dataset. We set $T \in \{1, 3\}$ for the metabolite dataset as due to its large dataset size.
% \baba{grid searchしているのは$T$だけ？$L$も？}
% \baba{over the validation set?}
We also set the dropout rate $0.2$ in Equations (\ref{internal_convolution}) and~(\ref{external_convolution}).
We use a validation set to tune those hyperparameters\footnote{Due to the limitation of the computational resources, we could not help limiting the hyperparameter ranges rather ``intuitively'' depending on the network sparsity and other data characteristics to perform thorough search over all possible hyperparameter combinations.}.

%%%%%%%%%%%%%%%%%%%%%%%%%%%%%%%%%%%
%%%%%%%%%%%%%%%%%%%%%%%%%%%%%%%%%%%

\subsection{Datasets and performance metric}\label{sec:datasets}

% added by Takigawa
% ---from here
We prepare three different chemical GoGs with different levels of sparsity and different weights of the tails of the degree distributions.
%The definition of edges between compounds differs from each network.
%\akita{データセットの並びを上から密→スパース→超スパースになるように変えてみました}
%\subsubsection{Drug--drug interaction network}

\subsubsection*{(a) Drug--drug interaction network} 
is a network of drug compounds where two compounds have an edge if they are known to interact, interfere, or cause adverse reactions when taken together.
From the DrugBank database\footnote{\url{https://www.drugbank.ca/releases/latest}}, we used $1{,}993$ approved drugs that have fewer than $64$ atoms. Out of all possible $\binom{1993}{2}=1,985,028$ compound pairs, $186{,}555$ have edges; the link density is $0.0940$ which means it is a relatively dense network.
Figure~\ref{fig:one} shows its degree distribution that shows a very heavy-tailed distribution.

% \color{red}
% For this dataset, we regard unlabeled links as negative links.
% We randomly choose n positive links and n negative links from possible pairs as a training dataset. As a test dataset, we randomly choose samples from the same distribution of the original network. The number of all test links is approximately $100{,}000$.
% \color{black}
%These edges are regarded as positive links, and others as negative ones.

% 妥当な設定?
% (他にも同じような設定でやっている人がいるなら、それを引用して、査読で突っ込まれないようにしたい).
We have only positive links in this dataset; this situation is sometimes dealt with positive-and-unlabeled learning~\cite{cerulo2010learning}; however, we just regard sampled no-links as the negative links for simplicity~\cite{doi:10.1093/bioinformatics/btn273}.
We randomly choose $n$ positive links
and $n$ no-links (i.e., negative links) as the training dataset.
We vary $n$ from $1$k to $10$k to investigate the importance of incorporating the information of the external graph by the external convolution.
% この後、$n$をいろいろ変えて実験する理由を言う
% (ネットワークの広範囲の情報を取り込める、つまりexternal convolutionの有用性を見るため)。
%The increase of training samples allows the model to
%consider the extensive information of the network;
%that is, by increasing training samples,
%we can evaluate the importance of incorporating the external convolution.
As the test dataset, we randomly extract positive and negative links from the same data distribution as the original network to preserve the data imbalance,
which results in $9{,}398$ positive links and $90{,}601$ negative links.
% 最後に具体的なデータ数を。

%\subsubsection{Drug--function network}
\subsubsection*{(b) Drug--function network} is a network of drug compounds where two compounds have an edge if they share the same target protein. From the original dataset~\cite{takigawa2011mining}, we used $3{,}918$ compounds that have fewer than $64$ atoms. Out of all possible $\binom{3918}{2}=7,673,403$ compound pairs, $35{,}562$ have edges; the link density is $0.0046$ which means it is a sparse network.
Figure~\ref{fig:two} shows its degree distribution that shows a relatively heavy-tailed distribution.
%These edges are regarded as positive links, and others as negative.

% \color{red}
% For this dataset, we follow the same procedure of~(i); the number of positive links is $1{,}390$ and negative links is $298{,}609$.
% \color{black}

As well as the drug-drug interaction dataset, this network also has only positive links;
therefore, we sample no-links as the negative links.
We have $1{,}390$ positive links and $298{,}609$ negative links in the test set.

%\subsubsection{Metabolite reaction network}
\subsubsection*{(c) Metabolite reaction network} is a network of metabolite compounds where two compounds have an edge if they are the substrate-product pair in an enzymatic reaction on metabolic pathways~\cite{kotera2014data}. In this study we collected $5{,}920$ compounds that have fewer than $64$ atoms. Out of all possible $\binom{5920}{2}=17,520,240$ compound pairs, only $5{,}041$ have edges; 
the link density is $0.0003$ which means it is an extremely sparse network.
%These edges are regarded as positive links, and other compound-compound pairs are regarded as negative links.
Figure~\ref{fig:three} shows its degree distribution that shows a light-tailed distribution.

% \color{red}
% For this dataset, we have explicitly labeled links, 5041 positive links and \color{green}$N$ (check later) \color{red} negative links and do not use any unlabeled links. Then, we follow the same procedure.
% The number of all test links is approximately $100{,}000$.
% \color{black}

Different from the other two datasets, this network has both $5,041$ positive links and $220,096$ negative links; the test set consists of $223$ positive links $9,777$ negative links.

%\harada{Negative labels are given in this data explicitly, so we handle those data as negative samples instead of handling rest of data as negative as Drug--drug and Drug--function.}

\begin{figure*}[tb]
  \begin{minipage}{\hsize}
 %\begin{center}
 \vspace{2mm}
   \includegraphics[width=\hsize]{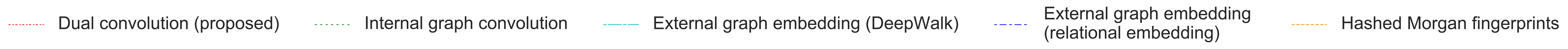}
 %  \end{center}
  \end{minipage}
 \end{figure*}
  
\begin{figure*}[tb]
\captionsetup[subfigure]{justification=centering}
\vspace{-2mm}
 \begin{minipage}{0.33\hsize}
  \centering
   \includegraphics[width=55mm]{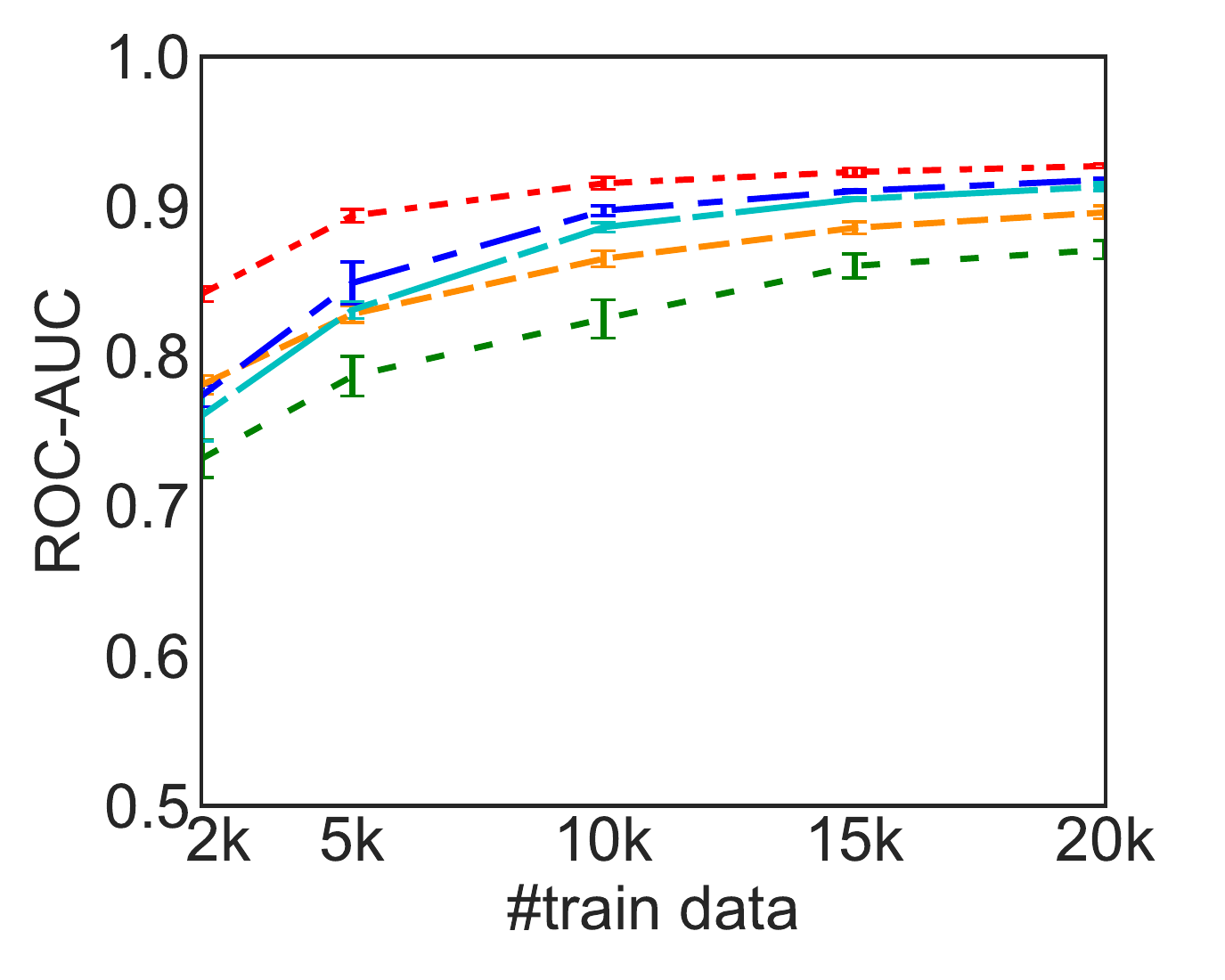}
   \label{fig:drug_roc}
%   \subcaption{Drug-drug interaction \\ (dense, extremely heavy-tailed)}
 \end{minipage}
 \begin{minipage}{0.33\hsize}
 \begin{center}
  \includegraphics[width=55mm]{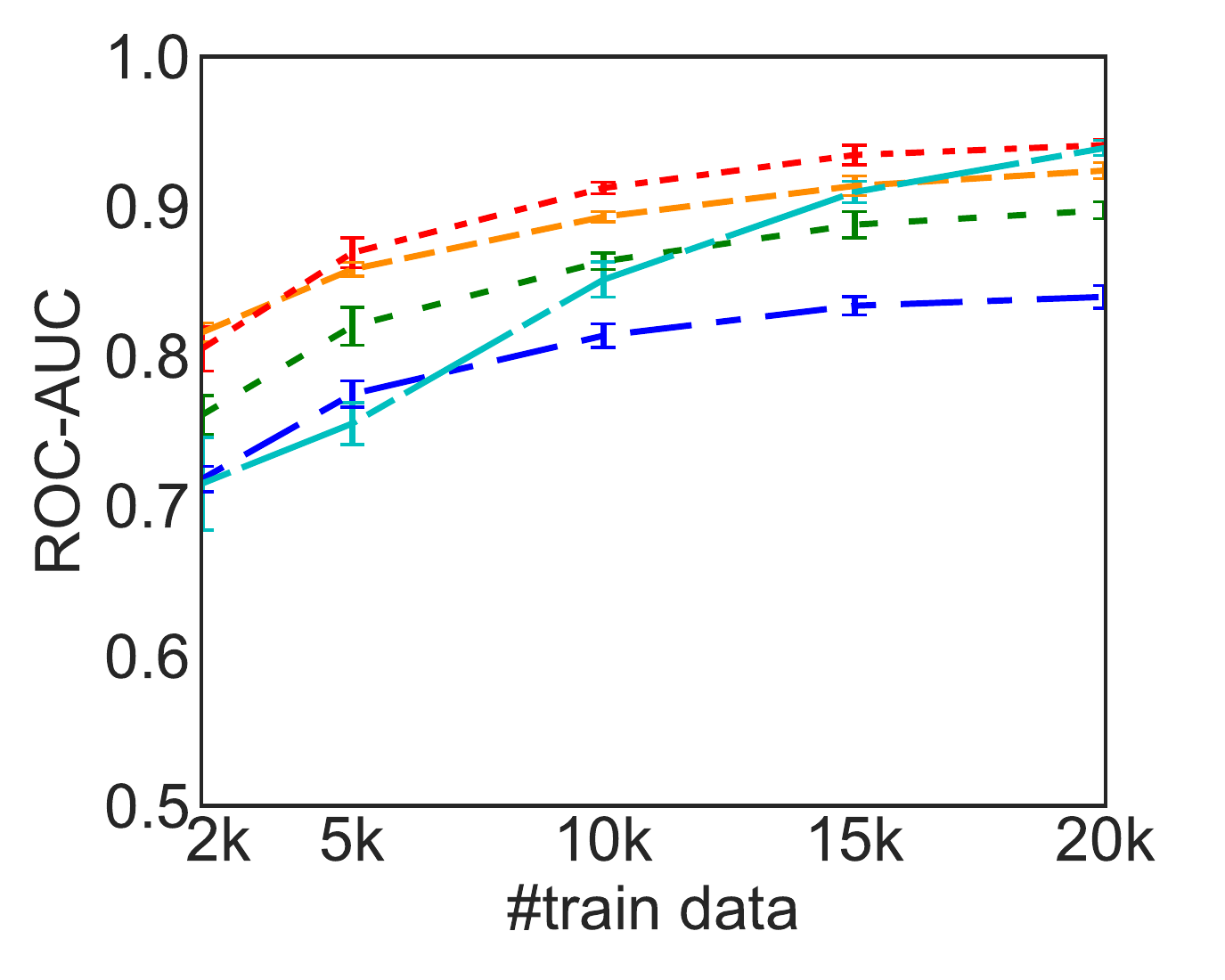}
 \end{center}
%   \subcaption{Drug-function \\ (sparse, heavy-tailed)}
  \label{fig:drugfunction_roc}
 \end{minipage}
 \begin{minipage}{0.33\hsize}
 \begin{center}
  \includegraphics[width=55mm]{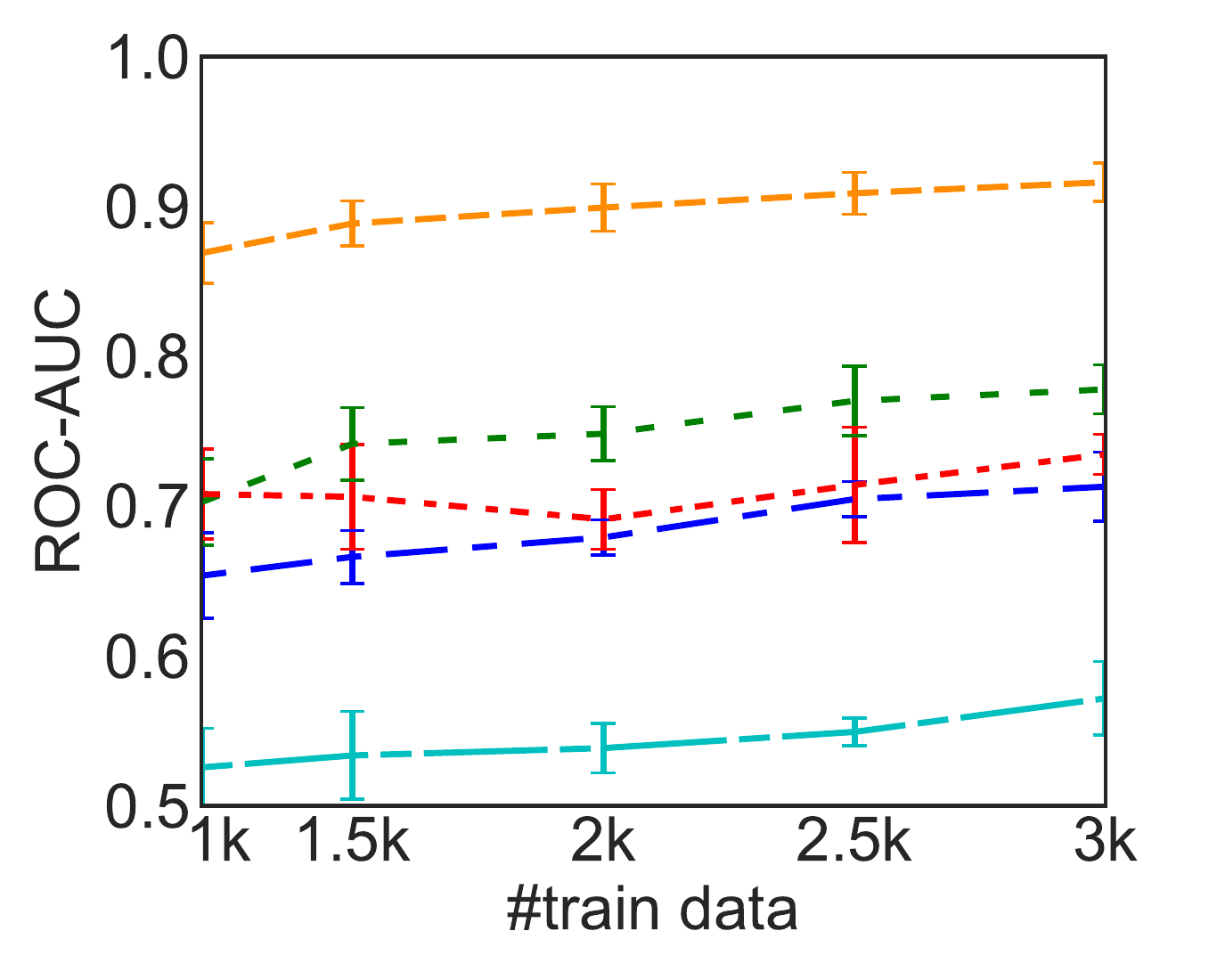}
 \end{center}
%   \subcaption{Metabolite \\ (extremely sparse, light-tailed)}
  \label{fig:metabolite_roc}
 \end{minipage}
% \caption{\small{\color{red}
% \label{fig:roc}
%  }}

 \captionsetup[subfigure]{justification=centering}
 \begin{minipage}{0.33\hsize}
  \centering
  \vspace{-3mm}
   \includegraphics[width=55mm]{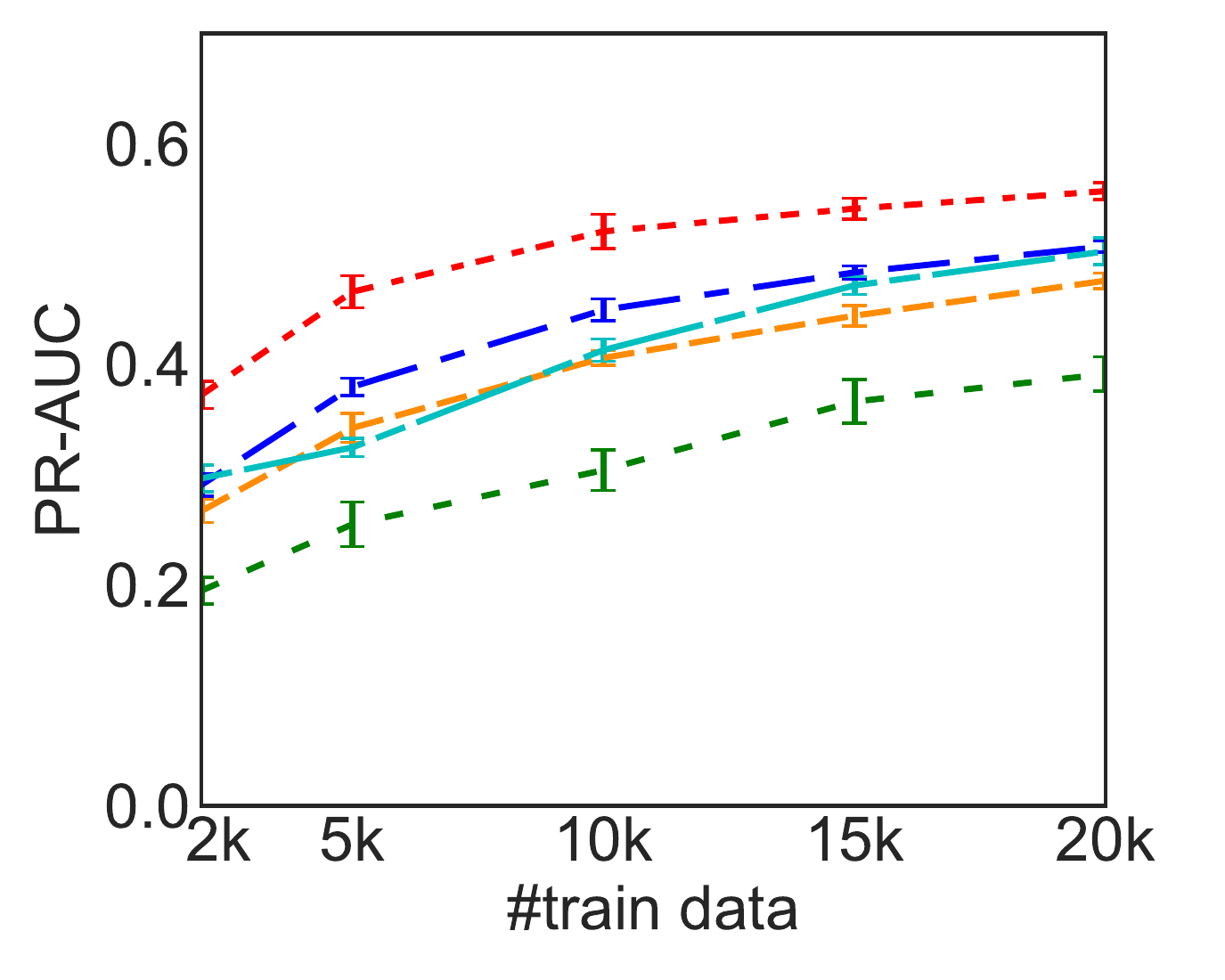}
  \subcaption{Drug-drug interaction \\ (dense, extremely heavy-tailed)}
  \label{fig:drug}
 \end{minipage}
 \begin{minipage}{0.33\hsize}
 \begin{center}
 \vspace{-3mm}
  \includegraphics[width=55mm]{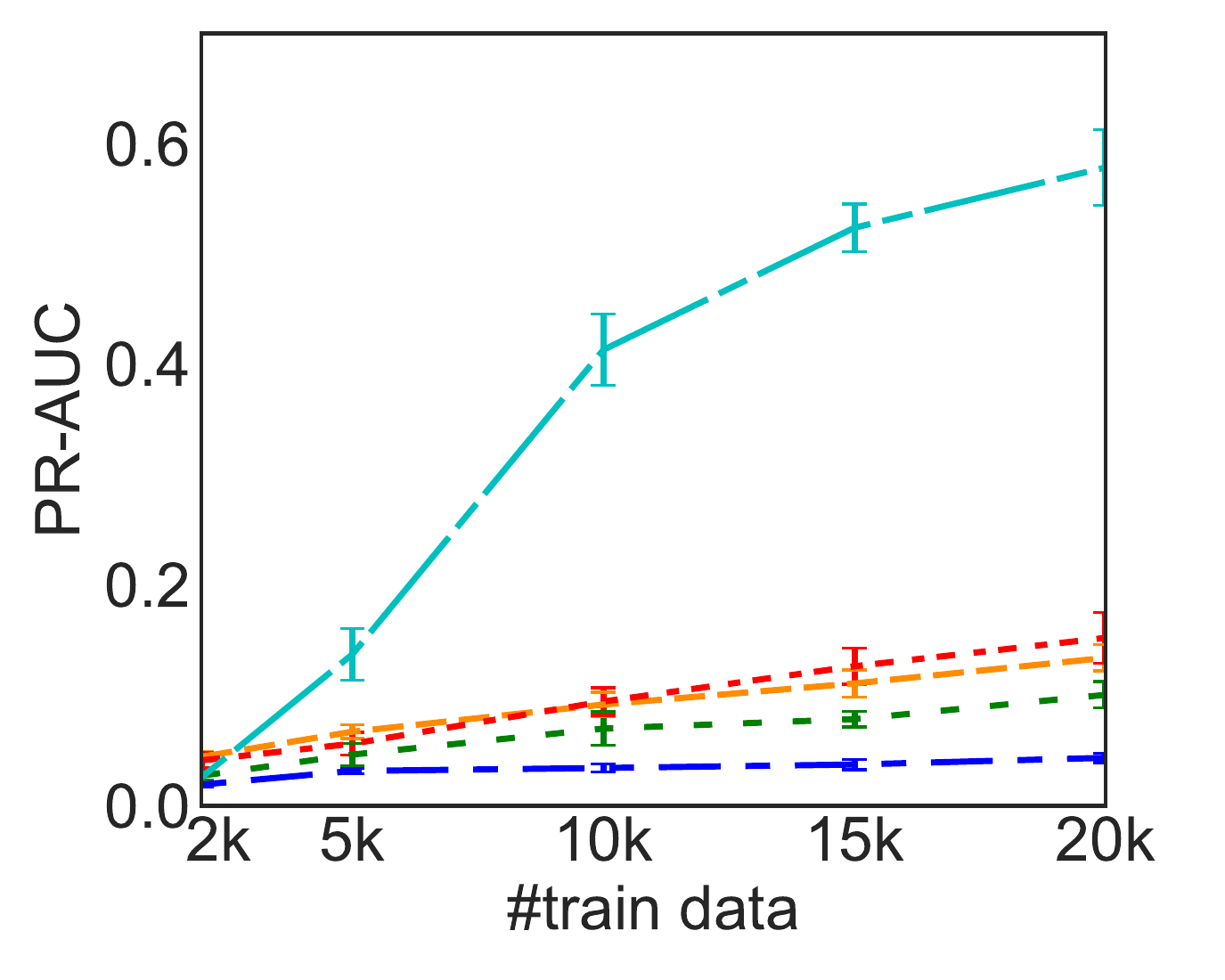}
 \end{center}
  \subcaption{Drug-function \\ (sparse, heavy-tailed)}
  \label{fig:drugfunction}
 \end{minipage}
 \begin{minipage}{0.33\hsize}
 \begin{center}
 \vspace{-3mm}
  \includegraphics[width=55mm]{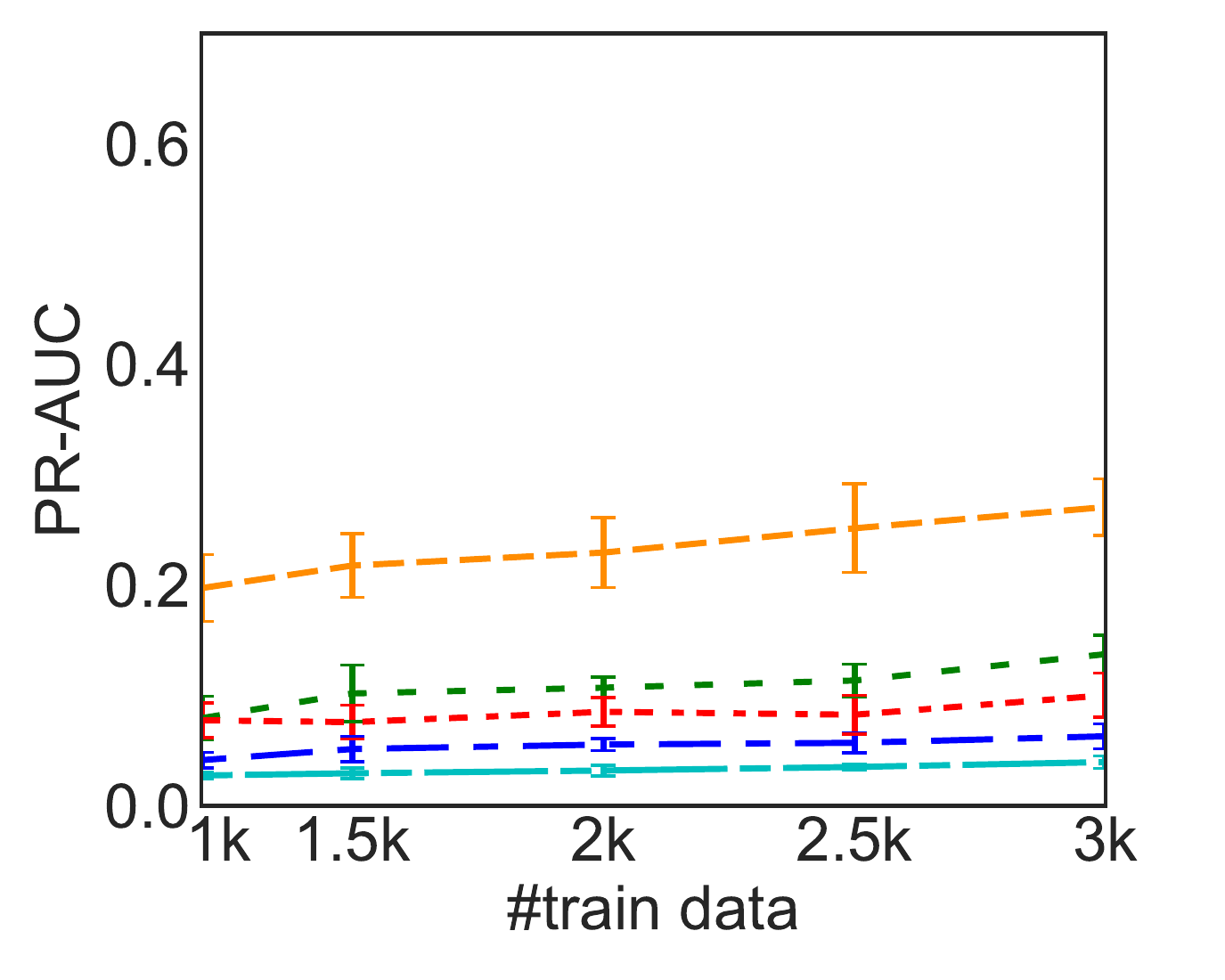}
 \end{center}
  \subcaption{Metabolite \\ (extremely sparse, light-tailed)}
  \label{fig:metabolite}
 \end{minipage}
\caption{\small{Link prediction performance in (a) the drug-drug interaction network, (b) the drug-function network, and (c) the metabolite reaction network.
  %\vspace{1mm} \\
The proposed method performs well for the first two networks ((a) and (b)) with heavy-tailed degree distributions.
On the other hand, in the extremely sparse and light-tailed external network (c), the external links are almost useless as features, and therefore the domain specific features (i.e., Morgan indices) perform the best. The internal convolution also suffers from the lack of external links as the training data. \color{black}
\label{fig:experiments}
 }}
\end{figure*}

\subsubsection*{}
In all of the three chemical networks, each external node is a relatively small compound (i.e. with up to $64$ atoms), while they have different levels of sparsity and degree distributions as shown in Fig.~\ref{fig:degree}, which is a typical variation in chemical networks. Our main interest is to obtain insights about the conditions of chemical networks in which the dual convolution is effective.
\iffalse
最後に、上の３つデータを使う意味を言う。
上のネットワークにおいて、内部グラフ（つまり化合物）にはそこまでバリエーションはない（原子数６４だし）。
一方で外部グラフ（つまりネットワーク）には、図に示すように、密度や次数に様々なバリエーションを持っている。
上のようないろいろなネットワークで、我々の手法と下に示す比較手法がどの程度動作するのか、適用範囲を考察する。
\fi
\color{black}

%%%%%%%%%%%%%%%%%%%%%%%%%%%%%%%%%%%%
%%%%%%%%%%%%%%%%%%%%%%%%%%%%%%%%%%%%

\subsection{Compared methods}
We compare the proposed dual convolutional network with several baselines, namely, 
(i) a model using only internal graph convolution,
(ii) models based only on external graph structures, and
(iii) a model based on hashed Morgan fingerprints instead of the internal graph convolution, %as internal (molecular) graph features and predict the link in the external graph.

\subsubsection*{(i) Internal graph convolution}
obtains $64$-dimensional representations of molecular graphs.
We do not use the external graph convolution, %and two feature vectors for two molecules created by only the internal convolution are directly input to the link prediction network.
but we create a feature vector for each molecule by the internal convolution and directly use it as an input to the link prediction network.
We use the same convolution formula as that by Duvenaud {\it et al.}~(\citeyear{NIPS2015_5954}).

\subsubsection*{(ii) External graph embedding}
%\kashima{文献を引いて、Externalな場合の標準的手法として位置づける}
%Only the pair of features on external graph\kashima{これなんでexternal convolutionって言わないの？（多少ムリしてでもいうべきでは）} \harada{L=0の時のexternal convolutionで、実際してないのであんまりよくないかなと・・・} }
is a standard approach to link prediction using only the external graph.
We test DeepWalk~\cite{perozzi2014deepwalk} that is one of the well-known embedding methods, and also test the general relational embedding model proposed by Yan {\it et al.}~\cite{yang2015embedding} where 
%\update{This is one of the most standard methods for external link prediction, and the well-known node2vec and DeepWalk are also variants of this method. }
the latent representation for each molecule is initialized to a $64$-dimensional random vector.
The link prediction network~\eqref{neural_network} is applied to a pair of molecules.

\subsubsection*{(iii) Hashed Morgan fingerprints}
define molecular graph features using chemical substructures.
We use $2048$-dimensional Morgan fingerprints as a feature vector of a molecule. The link prediction network \eqref{neural_network} is applied to a pair of molecules.
%In our experiment, we use $2048$-dimensional Morgan fingerprints because they showed better performance than using lower-dimensional fingerprints.

\if0
The regularization parameter, $\lambda$, was set to $0$ in the experiments.
\fi
%and the hyper-parameters and training details of neural network are as follows:
%the number of time steps $t$ (i.e., layers):
%1, 3, or 5 for drug-protein, 3, 5, or 8 or drug-drug, and 1 or 3 for chemical-interaction;
%optimization: ADAM \cite{a}, which is one of the SGD-based methods;
%batch size: 256 for drug-protein, 128 for drug-drug, 64 for chemical-interaction;
%regularization $\lambda$: 0, i.e., we do not consider the regularization.
%We perform grid search over a combination of the above hyper-parameters.

% \subsection{Results}

% \begin{itemize}
% \item 既存の特徴量（Morgan系）：一定の性能は出ているが、複数種の部分構造が同じインデックスに割り当てられていると考えられるので、
% 表現能力が他と比べて劣っていると思われる。
% \color{red}特徴ベクトルの次元数を増やして試すべきか\color{black}
% \item Only External：drug-drugのような密なネットワークに対しては高い性能を示しているが
% drug-proteinのような疎なネットワークでは性能が低下している。
% \item Only Internal: 内部構造の情報を取り入れるため、疎なネットワークに対しても頑強であると考えられる。密なdrug-drugではOnly Externalと比較すると性能としては劣っているが一定の性能は見られ、比較的疎なdrug-proteinではOnly Externalを上回る。
% \item external+internal(提案手法)：比較的密なネットワークではExternalの情報を取り入れて予測精度が向上する。
% 一方、最も疎なchemical-interactionでは外部構造が役に立たないのか、性能が低下する。
% \end{itemize}

\subsection{Results}

As we mentioned in Section~\ref{sec:datasets}, all the datasets have imbalance nature in terms of the number of positive and negative labels; therefore we measure the predictive performance of each method using (i)~ROC-AUC which is not affected by the label imbalance and (ii)~PR-AUC which can suitably evaluate the performance on imbalanced datasets.

Figure~\ref{fig:experiments} shows the comparison of the proposed method and four baselines with different training set sizes in terms of ROC-AUC and PR-AUC.
In Figure~\ref{fig:drug},
our dual convolution achieves consistently better ROC-AUC and PR-AUC scores over the baselines in the drug-drug interaction dataset.
This is probably due to the heavy-tailed degree distribution of its external graph.
In such networks, external links are likely to form a longer path, and therefore, the dual convolution successfully extracts structural features in the external graph.

\if0
\color{blue}
Figure~\ref{fig:drugfunction} shows the result for the drug-function network.
The advantage of the dual convolution still remains when the network is more sparse.
The external graph is moderately heavy-tailed and external links are still likely to form a path; the dual convolution benefits from both the internal and external graphs.
Note that this also allows DeepWalk to extract structural features from the external graph,
which is seen as the increase in performance as we increase the size of the training dataset.
\fi

Figure~\ref{fig:drugfunction} shows the result for the drug-function network.
It shows that the highest ROC-AUC score by the dual convolution.
The advantage of the dual convolution still remains when the external network is relatively sparse.
The external graph is moderately heavy-tailed and external links are still likely to form a path; the dual convolution benefits from both the internal and external graphs.
In addition, the ROC-AUC score of DeepWalk improves as the size of the training set increases; this implies that DeepWalk successfully extracts structural features from the relatively dense external graph.
Surprisingly, DeepWalk significantly outperform the others in terms of PR-AUC,
while our dual convolution achieves the best ROC-AUC score.
% 最後をこんな感じにしてみましたがどうでしょう？
Given that DeepWalk does not consider the internal graph structure at all,
information of the external graph is more crucial than the internal graphs in the drug-function network.

In contrast to the other two networks,
the metabolite network is an extremely sparse that has very few links
%while also having the largest number of nodes
and a light-tailed degree distribution.
The external links are almost useless in this network,
and therefore the relational embedding method and DeepWalk that solely depend on external links
perform poorly (Figure~\ref{fig:metabolite}).
Especially, DeepWalk performs the worst in terms of both ROC-AUC and PR-AUC because it cannot ``walk" over the external links.
Similarly, the proposed method cannot even benefit from the external convolution,
and it suffers from the sparse external links.
The lack of the external links as the training dataset
is also a severe limitation for extracting features from the internal graphs.
In such a sparse data domain, traditional off-the-shelf features
such as Morgan indices are still reliable choices.

In summary, our experimental results suggest that the dual convolution is effective for relatively dense networks, especially when both the internal and external structures must be considered in an integrated manner. 
Among the three networks, the links of the drug-drug interaction network represent direct chemical interactions between two compounds.
In such networks, nontrivial combination of different internal substructures of both ends of a link contributes to the interaction.
On the other hand, the links represent rather indirect chemical relations in the other two networks, where the benefit of the dual convolution remains limited.

\section{Related work}
%\kashima{????????????deep?n?????????K?o?`?????????????????????????????H?z???g???S???????????L???????????????C?????????A?|?C???g?|?C???g???????????????????c}
We briefly review the two-decade history of learning with graphs, especially from the viewpoints of internal graphs and external graphs.
Our present work attempts to unify these two lines of research that have been separately studied in different contexts.

%\subsection{Learning with internal graphs}
%\subsubsection{?????O???t???}?C?j???O}
The first generation of internally graph-structured data analysis appeared in the data-mining community.
The underlying idea is that local substructures of graphs are responsible for the properties of graphs.
Frequent pattern mining methods are extended to find such local substructures from a set of internally-structured graphs~\cite{inokuchi2000apriori,yan2002gspan}, and later are combined with boosting to find local patterns correlated with the target labels~\cite{kudo2005application}.

%\subsubsection{?J?[?l???@}
In contrast with the graph-mining approaches that explicitly find subgraph features, kernel methods perform implicit feature extraction.
The first graph kernels use paths as features~\cite{kashima2003marginalized,gartner2003graph}, and the recent state-of-the-art graph kernels use more complex subgraph patterns~\cite{shervashidze2011weisfeiler}.

%\subsubsection{?f?B?[?v???[?j???O}
Neural networks are also successfully applied to graphs, especially due to their capability of extracting flexible features from graphs.
Following the seminal work by Scarselli {\it et al.}~(\citeyear{scarselli2009graph}), various graph convolutional neural-network models have been proposed (e.g., \cite{NIPS2015_5954,niepert2016learning}).
Recent studies give a generalization of various existing models in terms of message passing~\cite{gilmer2017neural} and connections to the kernel methods~\cite{lei2017deriving}.

%\subsection{Learning with external graphs}
Analysis of external graph structured data is often called link mining~\cite{getoor2005link}. Its typical tasks include ranking, clustering, and classification of nodes, as well as link prediction.
Recently, node-embedding approaches that preserve node proximity in a graph have been extensively studied~\cite{perozzi2014deepwalk,grover2016node2vec}.
External graphs are also considered as object relations to which neural network approaches have also been applied successfully~\cite{socher2013reasoning,yang2015embedding}.

Because studies on graph-structured data analysis are wide-ranging and rapidly growing, it is quite difficult to cover all of them here; however, more to the point, internal and external graph analyses have been studied rather independently.
GoG is the very intersection where these two lines of studies meet, and our dual convolution approach makes it possible to extract features from both internal and external graph structures in an end-to-end manner.

\section{Conclusion}
% 本研究の目標であるinternal+external message passingを行った。
% 組み合わせだけを考慮するidが、疎なネットワークでは性能が低下している一方で
% 提案手法では一定の頑強性を示すことができた。
% %今後は、提案手法の性能をより適切に評価するために、テストデータの正負の割合や
% データセットの正負の割合を変更するなどして、どのようなデータセットに対して
% 最も性能を発揮するのか調査していく。

We addressed a problem of learning feature representations for a graph of graphs (GoG) and presented a dual convolution approach that combines both the external and internal graphs in an end-to-end manner. 
The proposed method was particularly designed for link prediction on an external graph and we demonstrated the effectiveness of the proposed method for predicting interactions among molecules by using three chemical network datasets.
Our dual convolution approach achieved high prediction performance even though the features were lower-dimensional compared to the off-the-shelf features in networks with heavy-tailed degree distributions. 
We found that the performance of the dual convolution approach becomes inferior on an extremely-sparse external network with a light-tailed degree distribution because of the difficulty of exploiting the information about the external network.  

Although we focused only on link prediction in this paper and the applications in other prediction tasks, such as node classification or clustering, will be addressed in future work.

%In this paper, we focused only on the link prediction task, because it is one of the most important GoG learning tasks; however we will study other applications including node classification and clustering.

\bibliographystyle{aaai}
\bibliography{reference}

\begin{thebibliography}{}

\bibitem[\protect\citeauthoryear{Cerulo, Elkan, and
  Ceccarelli}{2010}]{cerulo2010learning}
Cerulo, L.; Elkan, C.; and Ceccarelli, M.
\newblock 2010.
\newblock Learning gene regulatory networks from only positive and unlabeled
  data.
\newblock {\em BMC Bioinformatics} 11(1):228.

\bibitem[\protect\citeauthoryear{Duvenaud \bgroup et al\mbox.\egroup
  }{2015}]{NIPS2015_5954}
Duvenaud, D.~K.; Maclaurin, D.; Iparraguirre, J.; Bombarell, R.; Hirzel, T.;
  Aspuru-Guzik, A.; and Adams, R.~P.
\newblock 2015.
\newblock Convolutional networks on graphs for learning molecular fingerprints.
\newblock In {\em Advances in Neural Information Processing Systems (NIPS)}.

\bibitem[\protect\citeauthoryear{Gamo \bgroup et al\mbox.\egroup
  }{2010}]{gamo2010thousands}
Gamo, F.-J.; Sanz, L.~M.; Vidal, J.; de~Cozar, C.; Alvarez, E.; Lavandera,
  J.-L.; Vanderwall, D.~E.; Green, D. V.~S.; Kumar, V.; Hasan, S.; et~al.
\newblock 2010.
\newblock Thousands of chemical starting points for antimalarial lead
  identification.
\newblock {\em Nature} 465(7296):305--310.

\bibitem[\protect\citeauthoryear{G{\"a}rtner, Flach, and
  Wrobel}{2003}]{gartner2003graph}
G{\"a}rtner, T.; Flach, P.; and Wrobel, S.
\newblock 2003.
\newblock On graph kernels: hardness results and efficient alternatives.
\newblock In {\em Learning Theory and Kernel Machines}.
\newblock  129--143.

\bibitem[\protect\citeauthoryear{Getoor and Diehl}{2005}]{getoor2005link}
Getoor, L., and Diehl, C.~P.
\newblock 2005.
\newblock Link mining: a survey.
\newblock {\em SIGKDD Explorations} 7(2):3--12.

\bibitem[\protect\citeauthoryear{Gilmer \bgroup et al\mbox.\egroup
  }{2017}]{gilmer2017neural}
Gilmer, J.; Schoenholz, S.~S.; Riley, P.~F.; Vinyals, O.; and Dahl, G.~E.
\newblock 2017.
\newblock Neural message passing for quantum chemistry.
\newblock In {\em Proceedings of the 34th International Conference on Machine
  Learning (ICML)}.

\bibitem[\protect\citeauthoryear{Grover and
  Leskovec}{2016}]{grover2016node2vec}
Grover, A., and Leskovec, J.
\newblock 2016.
\newblock node2vec: Scalable feature learning for networks.
\newblock In {\em Proceedings of the 22nd ACM SIGKDD International Conference
  on Knowledge Discovery and Data Mining (KDD)},  855--864.

\bibitem[\protect\citeauthoryear{Inokuchi, Washio, and
  Motoda}{2000}]{inokuchi2000apriori}
Inokuchi, A.; Washio, T.; and Motoda, H.
\newblock 2000.
\newblock An apriori-based algorithm for mining frequent substructures from
  graph data.
\newblock In {\em Proceedings of the Fourth European Conference on Principles
  of Data Mining and Knowledge Discovery (PKDD)}.

\bibitem[\protect\citeauthoryear{Kashima, Tsuda, and
  Inokuchi}{2003}]{kashima2003marginalized}
Kashima, H.; Tsuda, K.; and Inokuchi, A.
\newblock 2003.
\newblock Marginalized kernels between labeled graphs.
\newblock In {\em Proceedings of the 20th International Conference on Machine
  Learning (ICML)}.

\bibitem[\protect\citeauthoryear{Kingma and Ba}{2015}]{kingma2014adam}
Kingma, D., and Ba, J.
\newblock 2015.
\newblock Adam: A method for stochastic optimization.
\newblock In {\em Proceedings of the Third International Conference for
  Learning Representations (ICLR)}.

\bibitem[\protect\citeauthoryear{Kotera \bgroup et al\mbox.\egroup
  }{2014}]{kotera2014data}
Kotera, M.; Tabei, Y.; Yamanishi, Y.; Muto, A.; Moriya, Y.; Tokimatsu, T.; and
  Goto, S.
\newblock 2014.
\newblock Metabolome-scale prediction of intermediate compounds in multistep
  metabolic pathways with a recursive supervised approach.
\newblock {\em Bioinformatics} 30(12):i165--i174.

\bibitem[\protect\citeauthoryear{Kudo, Maeda, and
  Matsumoto}{2005}]{kudo2005application}
Kudo, T.; Maeda, E.; and Matsumoto, Y.
\newblock 2005.
\newblock An application of boosting to graph classification.
\newblock In {\em Advances in Neural Information Processing Systems (NIPS)}.

\bibitem[\protect\citeauthoryear{Lei \bgroup et al\mbox.\egroup
  }{2017}]{lei2017deriving}
Lei, T.; Jin, W.; Barzilay, R.; and Jaakkola, T.
\newblock 2017.
\newblock Deriving neural architectures from sequence and graph kernels.
\newblock In {\em Proceedings of the 34th International Conference on Machine
  Learning (ICML)}.

\bibitem[\protect\citeauthoryear{Lounkine \bgroup et al\mbox.\egroup
  }{2012}]{lounkine2012large}
Lounkine, E.; Keiser, M.~J.; Whitebread, S.; Mikhailov, D.; Hamon, J.; Jenkins,
  J.~L.; Lavan, P.; Weber, E.; Doak, A.~K.; C{\^o}t{\'e}, S.; et~al.
\newblock 2012.
\newblock Large-scale prediction and testing of drug activity on side-effect
  targets.
\newblock {\em Nature} 486(7403):361--367.

\bibitem[\protect\citeauthoryear{Medina-Franco \bgroup et al\mbox.\egroup
  }{2013}]{medina2013shifting}
Medina-Franco, J.~L.; Giulianotti, M.~A.; Welmaker, G.~S.; and Houghten, R.~A.
\newblock 2013.
\newblock Shifting from the single to the multitarget paradigm in drug
  discovery.
\newblock {\em Drug Discovery Today} 18(9):495--501.

\bibitem[\protect\citeauthoryear{Mordelet and
  Vert}{2008}]{doi:10.1093/bioinformatics/btn273}
Mordelet, F., and Vert, J.-P.
\newblock 2008.
\newblock Sirene: supervised inference of regulatory networks.
\newblock {\em Bioinformatics} 24(16):i76--i82.

\bibitem[\protect\citeauthoryear{Morgan}{1965}]{morgan1965generation}
Morgan, H.~L.
\newblock 1965.
\newblock The generation of a unique machine description for chemical
  structures-a technique developed at chemical abstracts service.
\newblock {\em Journal of Chemical Documentation} 5(2):107--113.

\bibitem[\protect\citeauthoryear{Niepert, Ahmed, and
  Kutzkov}{2016}]{niepert2016learning}
Niepert, M.; Ahmed, M.; and Kutzkov, K.
\newblock 2016.
\newblock Learning convolutional neural networks for graphs.
\newblock In {\em Proceedings of the 33rd International Conference on Machine
  Learning (ICML)},  2014--2023.

\bibitem[\protect\citeauthoryear{Perozzi, Al-Rfou, and
  Skiena}{2014}]{perozzi2014deepwalk}
Perozzi, B.; Al-Rfou, R.; and Skiena, S.
\newblock 2014.
\newblock Deep{W}alk: online learning of social representations.
\newblock In {\em Proceedings of the 20th ACM SIGKDD International Conference
  on Knowledge Discovery and Data Mining (KDD)},  701--710.

\bibitem[\protect\citeauthoryear{Scarselli \bgroup et al\mbox.\egroup
  }{2009}]{scarselli2009graph}
Scarselli, F.; Gori, M.; Tsoi, A.~C.; Hagenbuchner, M.; and Monfardini, G.
\newblock 2009.
\newblock The graph neural network model.
\newblock {\em IEEE Transactions on Neural Networks} 20(1):61--80.

\bibitem[\protect\citeauthoryear{Shervashidze \bgroup et al\mbox.\egroup
  }{2011}]{shervashidze2011weisfeiler}
Shervashidze, N.; Schweitzer, P.; Leeuwen, E. J.~v.; Mehlhorn, K.; and
  Borgwardt, K.~M.
\newblock 2011.
\newblock Weisfeiler-{L}ehman graph kernels.
\newblock {\em Journal of Machine Learning Research} 12(Sep):2539--2561.

\bibitem[\protect\citeauthoryear{Socher \bgroup et al\mbox.\egroup
  }{2013}]{socher2013reasoning}
Socher, R.; Chen, D.; Manning, C.~D.; and Ng, A.
\newblock 2013.
\newblock Reasoning with neural tensor networks for knowledge base completion.
\newblock In {\em Advances in Neural Information Processing Systems (NIPS)}.

\bibitem[\protect\citeauthoryear{Sutskever, Vinyals, and
  Le}{2014}]{sutskever2014sequence}
Sutskever, I.; Vinyals, O.; and Le, Q.~V.
\newblock 2014.
\newblock Sequence to sequence learning with neural networks.
\newblock In {\em Advances in Neural Information Processing Systems (NIPS)}.

\bibitem[\protect\citeauthoryear{Takigawa, Tsuda, and
  Mamitsuka}{2011}]{takigawa2011mining}
Takigawa, I.; Tsuda, K.; and Mamitsuka, H.
\newblock 2011.
\newblock Mining significant substructure pairs for interpreting
  polypharmacology in drug-target network.
\newblock {\em PLoS ONE} 6(2):e16999.

\bibitem[\protect\citeauthoryear{Tokui, Oono, and
  Hido}{2015}]{tokui2015chainer}
Tokui, S.; Oono, K.; and Hido, S.
\newblock 2015.
\newblock Chainer: a next-generation open source framework for deep learning.
\newblock In {\em Proceedings of Workshop on Machine Learning Systems at NIPS
  2015}.

\bibitem[\protect\citeauthoryear{Yan and Han}{2002}]{yan2002gspan}
Yan, X., and Han, J.
\newblock 2002.
\newblock g{S}pan: Graph-based substructure pattern mining.
\newblock In {\em Proceedings of the Second IEEE International Conference on
  Data Mining (ICDM)}.

\bibitem[\protect\citeauthoryear{Yang \bgroup et al\mbox.\egroup
  }{2015}]{yang2015embedding}
Yang, B.; Yih, W.-t.; He, X.; Gao, J.; and Deng, L.
\newblock 2015.
\newblock Embedding entities and relations for learning and inference in
  knowledge bases.
\newblock In {\em Proceedings of the Third International Conference on Learning
  Representations (ICLR)}.

\end{thebibliography}

\end{document}